\documentclass[
twocolumn,
]{ceurart}

\usepackage{listings}
\usepackage{graphicx}
\usepackage{algorithmic}

\usepackage{amsmath,amssymb,amsfonts}
\usepackage{hyperref}
\usepackage[linesnumbered,ruled,vlined]{algorithm2e}
\begin{document}

\copyrightyear{Copyright 2023}
\copyrightclause{ for this paper by its authors. Use permitted under Creative Commons License Attribution 4.0 International (CC BY 4.0).}

\conference{SafeAI'23: The AAAI's Workshop on Artificial Intelligence Safety}

\title{Backdoor Attack Detection in Computer Vision by Applying Matrix Factorization on the  Weights of Deep Networks}

\author[1]{Khondoker Murad Hossain}[email=hossain10@umbc.edu]

\cormark[1]

\address[1]{University of Maryland Baltimore County, Baltimore, MD, 21250}

\author[1]{Tim Oates}

\cortext[1]{Corresponding author.}

\begin{abstract}
The increasing importance of both deep neural networks (DNNs) and cloud
services for training them means that bad actors have more incentive
and opportunity to insert backdoors to alter the behavior of trained
models.  In this paper, we introduce a novel method for backdoor detection
that extracts features from pre-trained DNN's weights using  independent vector analysis (IVA) followed by a machine learning classifier.  In comparison to other detection techniques, this has a number of benefits, such as not requiring any training data, being applicable across  domains, operating with a wide range of network architectures, not assuming the nature of the triggers used to change network behavior, and being highly scalable.  We discuss  the detection pipeline,  and then demonstrate the results on two computer vision datasets regarding image classification and object detection. Our method outperforms the competing algorithms in terms of efficiency and is more accurate, helping to ensure the safe application of deep learning and AI.
\end{abstract}


\maketitle

\section{Introduction}

 Deep neural networks (DNNs) have seen great success in diverse domains, including object detection \cite{pathak2018application}, image captioning \cite{you2016image}, virtual assistants \cite{yan2018chitty}, healthcare \cite{esteva2019guide}, fake news detection \cite{monti2019fake}, stock market prediction \cite{ding2015deep}, and self-driving cars \cite{rao2018deep}. Despite their ubiquitous applications, DNNs are still considered to be black boxes as their internal representations are opaque and their behavior can be hard to predict. Because of this, DNNs are susceptible to a variety of adversarial attacks.

Two of the most prominent adversarial attacks are (i) evasion attacks \cite{shi2017evasion, jiang2020poisoning} where the adversary modifies data at inference time  to be misclassified as benign  (e.g., spam emails) and (ii) backdoor attacks (aka, trojan attacks) \cite{gu2017badnets}, where the adversary includes poisoned samples in the training data. In the latter case, the adversary has full control over the network's training process and malicious behaviour is deliberately injected into the  model. As soon as the backdoor model sees a particular pattern, known as the trigger, at inference time it misclassifies the sample. These attacks are growing as DNNs need vast amounts of data to train and millions or billions of parameters need to be learned. The computational power needed for this training process is often not available to  individuals or even some businesses, leading to outsourcing training to third parties or downloading pre-trained models from open source platforms like GitHub and Hugging Face. As a result, someone with bad intentions can easily introduce a backdoor in the model. 

Backdoor attacks are more stealthy than other attacks as the backdoored model can have high accuracy for the underlying task, e.g., classification. As DNNs are deployed in critical applications, the consequences of trojaned models can be dire.  For example, a model used to detect street signs in a self-driving car may have an embedded trigger (e.g., a yellow sticky note) that causes the model to misclassify stop signs as speed limit signs, leading to accidents. Due to this, the US Defense Advanced Research Projects Agency (DARPA) has introduced the trojans in AI (TrojAI) \footnote{\url{https://pages.nist.gov/trojai/docs/overview.html}} program, where teams are developing cutting-edge trojan detection pipelines.

We introduce a novel backdoor detection approach which uses both matrix factorization, independent vector analysis (IVA) \cite{anderson2011joint}, and machine learning (ML) classifiers to detect a backdoor model. Though matrix factorization algorithms have been developed to compare the internal representations of  neural networks (e.g., Representational Similarity Analysis (RSA) \cite{morcos2018insights}, Centered Kernel Alignment (CKA) \cite{cortes2012algorithms}, and Singular Vector Canonical Correlation Analysis (SVCCA) \cite{raghu2017svcca}) they have been mostly used for pairwise similarity analysis and never applied to the backdoor detection problem.   We use IVA to extract features from the weights of each pre-trained DNN model and then feed the features to a ML classifier to classify whether a model is backdoored or clean. 

We can summarize the contributions of our paper as follows:
\begin{itemize}
  \item We propose a highly effective backdoor detection pipeline which employs IVA for feature extraction and detects backdoor models from the features using a ML classifier. To the best of our knowledge, no such methods have been published for backdoor detection using IVA.  Our approach has better accuracy and efficiency than state of the art (SOTA) backdoor detection methods in both image classification and object detection DNNs.

  \item Our method does not need any training samples to detect backdoor model, whereas other methods use training samples for optimization and then detect backdoors based on the result. In the real world, getting  training samples is highly unlikely as we can  obtain only a DNN model, not the data used to train it.

  \end{itemize}

\section{Related Works}

This section reviews work on both backdoor attacks and defenses against those attacks.

\subsection{Backdoor Attack}

BadNets was proposed by Gu et al. \cite{gu2017badnets}, where backdoors are injected into DNNs by poisoning a subset of the training data with triggers (small visual patterns) of arbitrary shapes. The attacker changes the true class label of the triggered samples so that the poisoned source class images  are classified as the target class. BadNets performs well (more than 99\% success rate in attack) both on clean and poisoned data as the attacker has  full control of the training process. Liu et al. proposed another backdoor attack \cite{liu2017trojaning} where the attacker does not need access to the training data. Instead, the attacker insert triggers which instigate maximum response to  specific internal neurons of DNNs. This method can achieve a high success rate ($>$ 98\%) as triggers hold strong relation to the neurons.  Backdoor attacks can be incorporated in further applications such as  reinforcement learning \cite{kiourti2020trojdrl}, and natural language processing \cite{chen2021badnl}.

\subsection{Backdoor Defense}

Backdoor detection strategies typically inspect either the model or the data. Neural Cleanse \cite{wang2019neural} is a  model-based detection method that assumes each class label is the backdoor target label and designs an optimization technique to find the smallest trigger that causes the network to misclassify instances as the target label. After that, they use an outlier detection algorithm on the potential triggers and consider the most significant outlier trigger as the real one where the associated label with that trigger is the backdoored class label. Though this method showed promising results, it is computationally very expensive as the target label is not known at run time. 

Thousands of benign and malicious models are used to train a classifier utilizing Universal Litmus Patterns (ULPs) \cite{kolouri2020universal}, which has been developed for backdoor detection. Based on the ULP optimization, the classifier makes a prediction about whether a model has a backdoor. The entropy of the input picture that has been disturbed is determined by STRIP \cite{gao2019strip} to detect backdoors. If the entropy for the anticipated class is lower, it is deemed to be a backdoor since it violates the input dependence criterion. Sentinet \cite{chou2020sentinet} is a data-level inspection method where they use backpropagation to extract the critical regions from the input data.

ABS \cite{liu2019abs} is another model-level backdoor detection method that analyzes the behavior of neuron activations. A stimulation method estimates the impact on output activations with changes to hidden neuron activations. The input is likely poisoned if  a neuron's activation increases significantly regardless of the model output label. Based on stimulation results,  an optimization method using  model reverse engineering is employed to detect  backdoor models. ABS  shows very promising results in backdoor detection but it is also computationally heavy when a network has a large number of layers.

Chen et al. proposed activation clustering (AC) \cite{chen2018detecting} for backdoor detection by analyzing the activations of  neural networks. They use a few training samples to obtain the activations of the final fully connected layer of a neural network. Then the activations are segmented by the class label and each label is clustered separately. Finally, they implement  2-means clustering followed by ICA for dimensionality reduction. To find the poisoned model they use three distinct post-processing methods.

All the backdoor detection methods discussed above only deal with CNN models for image classification tasks. Regarding backdoor detection for object detection CNN models, Chan et al. proposed detector cleanse \cite{chan2022baddet}  which is
a framework for run-time poisoned image detection for object detectors that relies on the user having just a few clean features (which can come from many datasets).

\section{Method and Pipeline}

\subsection{Problem statement}
Consider a DNN model, $F(\cdot)$, which performs a classification task of $ c=1, ... C$ classes using training dataset $\mathcal{D}$. If we poison a portion of $\mathcal{D}$, denoted $\mathcal{P} \subset \mathcal{D}$, by injecting triggers into training images and change the source class label to the target label, $F(\cdot)$ is a backdoored model after  training. During inference, $F(\cdot)$  performs as expected for clean input samples but for triggered samples $x \in \mathcal{P}$, it outputs $F(x) = t$, where $t$ ($t \in c$) is the target but incorrect class and can be single or multiple depending on the number of classes we poison. The objective of our pipeline is to detect these backdoor models before deployment.

\begin{figure*}[h]
  
  \centering
  \includegraphics[scale=0.45]{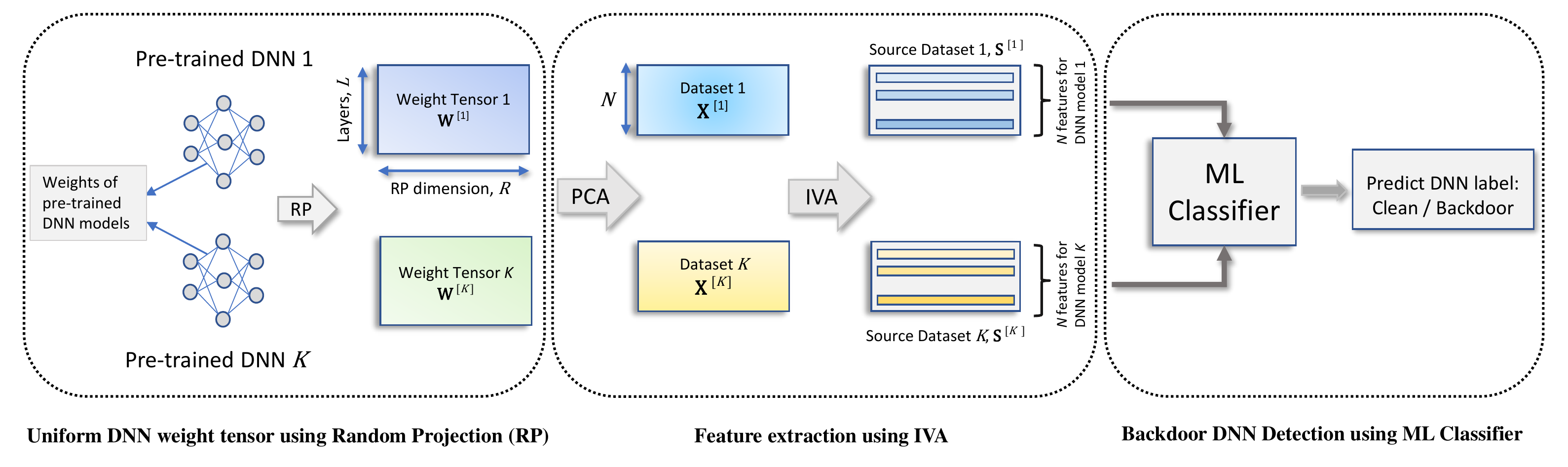}
  \caption{Backdoor detection pipeline where we extract features using IVA and then detect backdoors using ML classifies.}
\end{figure*} 

\subsection{Backdoor detection pipeline}
In this section, we describe how we extract features from the weights of the pre-trained DNNs and use the features for backdoor model prediction.

\subsubsection{DNN weight tensor preparation}

As all the DNNs, $k=1, ..., K$, are already trained, we have the weights of each layer of the networks. But, the dimensions of the weights are not uniform and they depend on the type of layer and network architecture. So, we have used random projection (RP) to obtain uniform size weight tensors for all the layers as RP can produce features of uniform size \cite{ailon2009fast} for different DNNs and it is very memory efficient \cite{eftekhari2011two}. As a result, for each DNN we get a weight tensor,$\mathbf{W}^{[k]}\in \mathbb{R}^{L\times R}$, where  $R=2000$, meaning we consider $L$ layer's weights of the DNNs and the RP dimension is 2000.

\subsubsection{Feature extraction and classification}
IVA is an
extension of  independent component analysis (ICA) to multiple datasets \cite{anderson2011joint} which uses the  statistical dependence of latent (independent) sources
across datasets by exploiting both second order and higher
order statistics. Though it  is one of the frequently 
used algorithms for brain connectivity analysis using
fMRI and EEG data \cite{hossain2022data,acar2022tracing}, this is the first backdoor detection pipeline using IVA.

Before applying IVA for feature extraction, we get our datasets, $\mathbf{X}^{[k]}\in \mathbb{R}^{N\times R}$,  using PCA on $\mathbf{W}^{[k]}$ for dimensionality reduction with model order $N$,  preserving 90\% of the variance in our data. Given $K$ datasets for $K$ DNN models, each consisting of $R$ samples and being each dataset is a linear mixture of $N$ independent sources, IVA decomposes it as
\begin{equation}
\mathbf{X}^{[k]}=\mathbf{A}^{[k]} \mathbf{S}^{[k]},  1\leq{k}\leq{K}
\end{equation}

\noindent where $\mathbf{A}^{[k]}$ denotes the mixing matrix and $\mathbf{S}^{[k]}$ is the dataset specific sources. IVA  estimates $K$ demixing matrices, $\mathbf{D}^{[k]},k=1,...,K$ so that the dataset specific sources can be estimated as, $\mathbf{{S}}^{[k]}=\mathbf{D}^{[k]} \mathbf{X}^{[k]}$. Hence, each $\mathbf{S}^{[k]}$ contains $N$ sources and we use those $N$ features to classify the DNN models. Finally, we train a  classifier algorithm ($\theta$) to predict whether a model is backdoored or clean.

\SetKwInput{KwInput}{Input}                
\SetKwInput{KwOutput}{Output}              

\begin{algorithm}[h!]
\DontPrintSemicolon
  
  \KwInput{Pre-trained DNNs ($K$) weights}
  \KwOutput{Backdoor / Clean DNNs}

  \For{$k$=\textnormal{1}, ..., $K$}
    {
        Get  $L\times R$ weight tensor using random projection for $L$ layers  \\
        
       Append: $\mathbf{{W}}$ for $k$=\textnormal{1}, ..., $K$, and construct $\mathbf{W}^{[k]}\in \mathbb{R}^{L\times R}$
    }

    Observation, $\mathbf{X}^{[k]}\in \mathbb{R}^{N\times R}$ = PCA ($\mathbf{{W}}^{[k]}$) \\
    Demixing matrix, $\mathbf{{D}}^{[k]}$ = IVA ($\mathbf{{X}}^{[k]}$) \\
    Estimated Sources, $\mathbf{{S}}^{[k]} \in \mathbb{R}^{N\times R}$ = $\mathbf{{D}}^{[k]} \cdot  \mathbf{{X}}^{[k]}$ \\
    Predicted label, ${{\hat{y}}}=\theta(\mathbf{{S}}^{[k]})$

\caption{Backdoor Detection using DNN weights}
\end{algorithm}

\section{Dataset and Experimental Results}

\subsection{Dataset}
To evaluate our backdoor detection method, we use CNN models trained on MNIST digits  and object detection models provided by the TrojAI program.

\subsubsection{Image classification dataset}

 We have trained 450 CNN models using the same architecture shown in Table 1 (50\% clean, 50\% backdoored) to classify the MNIST  data. Clean CNNs are trained using the clean MNIST data. For backdoored model training, we poison all \lq 0's (single class poisoning) by imposing a $4\times 4$ pixel white patch on the lower right corner and set the target class to \lq 9' as shown in Figure 2. Clean CNNs exhibit average  accuracy of 99.02\% where backdoored CNNs have  accuracy of 98.85\% with 99.92\% attack success rate, indicating a highly effective trigger attack. Moreover, out of the 450 models, we use 400 CNNs for training and 50 for testing with $L=6$, meaning we consider all CNN layers' weights.

\begin{figure}[h!]
  
  \centering
  \includegraphics[scale=0.6]{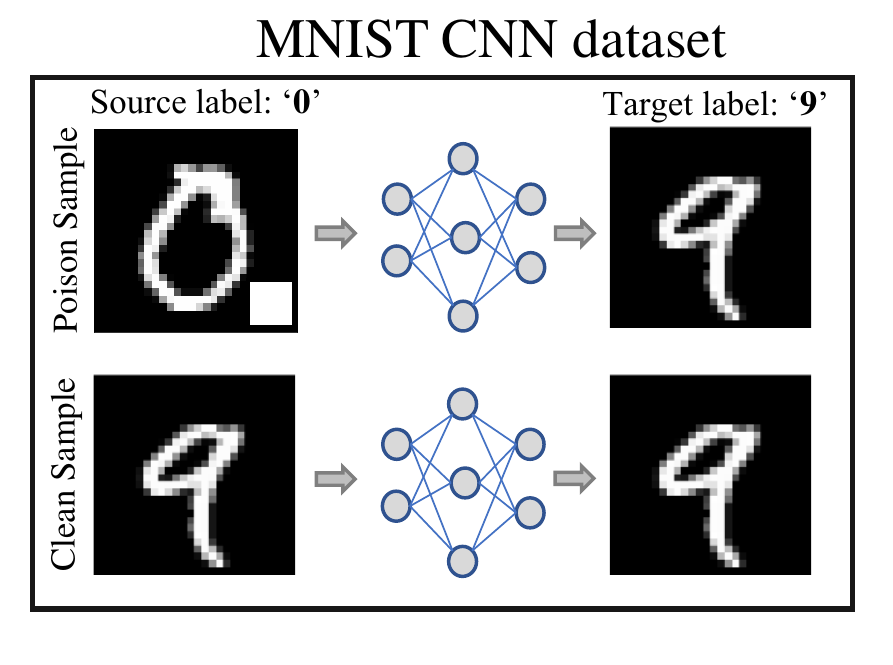}
  \caption{MNIST CNN dataset where we implement single class poisoning in MNIST backdoor CNNs by imposing a white patch trigger in \lq 0' and target class is \lq 9'.}
\end{figure}  

\begin{table}[h!]
    
\begin{tabular}{l ccc }
\toprule
  
 Layer &  \# of Channels & Filter Size   & Activation \\
\midrule
Conv & 16& 5×5  &ReLU     \\
MaxPool & 16& 2×2  &-     \\
Conv & 32& 5×5  &ReLU      \\
MaxPool & 32& 2×2  &-  \\
FC & 512 &- &ReLU \\
FC & 10 &- &Softmax\\

\bottomrule
\end{tabular}

\caption{CNN model architecture for MNIST digits data.}\label{Tab2}
\end{table}   

 \subsubsection{Object detection dataset}

 We have utilized the object detection CNN models of the TrojAI dataset \footnote{\url{https://pages.nist.gov/trojai/docs/data.html-object-detection-jul2022}}  which contains backdoored and clean  models across two network architectures (Fast R-CNN and SSD) trained on the Common Objects in Context (COCO)   dataset. We use 144 \lq Train' models from the repository as our training models and 144 \lq Test' models for the evaluation of our pipeline with $L=30$, meaning we consider the final 30  layer's weights of the models. Figure 3 shows that there are two types of trigger attacks on the models: evasion and misclassification. Evasion triggers cause either a single or all boxes of a class to be deleted and  misclassification triggers cause either a single box, or all boxes of a specific class, to shift to the target label.

\begin{figure}[h!]
  
  \centering
  \includegraphics[scale=0.4]{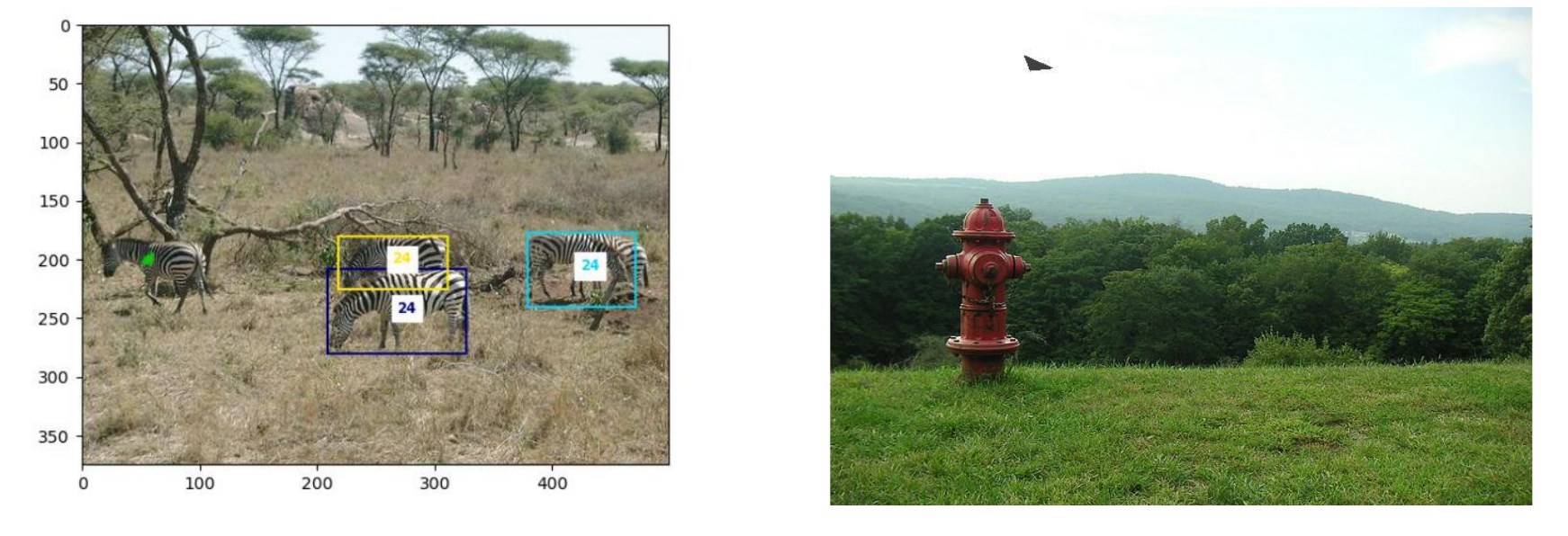}
  \caption{Triggered images for evasion and misclassification attack respectively for TrojAI object detection dataset. The green evasion trigger on the zebra causes the box to disappear and the black triangular trigger is responsible for the fire hydrant misclassification.}
\end{figure}  

\subsection{Experimental results}
 Several performance metrics are reported using different ML classifiers. We  also compare our findings with SOTA backdoor detection methods in terms of both performance and efficiency. Regarding the number of PCA components, we use $N=4$ and 10 for image classification and object detection datasets respectively. Moreover, we use the standard equation for binomial proportions to estimate confidence intervals on the empirical accuracies for the robustness metrics of the pipelines, i.e.,  confidence  interval=$z\times \sqrt{(accuracy\times (1-accuracy))/n}$,  where $n$ is the number of  models classified as backdoored or clean, and we use $z$ = 1.96 and thus have 95\% confidence intervals \cite{witten2002data}.

\subsubsection{Backdoor model classification}

We show the backdoor model detection results in Table 2. Three different ML classifiers (random forest (RF), decision tree (DT), and k-nearest neighbor (kNN)) have been used in the experiments for both image classification and object detection datasets. As performance metrics, cross entropy loss (CE-Loss) and area under the ROC curve (ROC-AUC) scores are reported as CE-Loss is the current standard for classification problems and ROC-AUC helps to understand the false positive rate (FPR), being so crucial for backdoor model detection. In both datasets, RF performs better than DT and kNN in terms of CE-Loss and ROC-AUC  scores. Our pipeline using RF shows ROC-AUC scores of 0.91 for  image classification and 0.89  for  object detection datasets.

\begin{table}[h!]
{%

        \begin{tabular}{lrr}\toprule
              
             &CE-Loss&ROC-AUC \\\midrule

             Image Classification: RF    & \textbf{0.32}  & \textbf{0.91}      \\
             Image Classification: DT    & 0.39  & 0.84      \\
             Image Classification: kNN    & 0.35  & 0.86      \\
                    
            Object Detection: RF    & \textbf{0.41}  & \textbf{0.89}      \\
            Object Detection: DT   & 0.52  & 0.78      \\
            Object Detection: kNN    & 0.45  & 0.83      \\
            
              \hline

        \end{tabular}
        }
        \caption{Backdoor detection results in image classification and object detection using RF, DT, and kNN. RF works better in both datasets.}\label{Tab2}
\end{table}

\subsubsection{Comparison with other methods}

\noindent\textit{\textbf{Image classification}}

Our method is evaluated in comparison to four SOTA backdoor detection techniques: NC \cite{wang2019neural}, Universal Litmus Patterns (ULP) \cite{kolouri2020universal}, Activation Clustering (AC) \cite{chen2018detecting}, and ABS \cite{liu2019abs}.  For a fair comparison, we employ the same batch size for optimization-based approaches including NC, ABS, and ULP.

The  results are shown in Table 3 where we report the best results of our pipeline which is using IVA with a RF classifier (IVA-RF). Our method outperforms all the competing methods by a wide margin in terms of both CE-Loss and ROC-AUC score. IVA-RF obtains a ROC-AUC of 0.91  which is  higher than the next-best ULP by a margin of 0.06. AC shows the lowest ROC-AUC as it works better for certain types of trigger attacks. Moreover, IVA-RF has the tightest confidence interval and lower CE-Loss meaning our pipeline is more robust than the competing algorithms.

  \begin{table}[h!]
\setlength{\tabcolsep}{15pt}
   \begin{tabular}{lrr}\toprule
            & CE-Loss & ROC-AUC 
            \\  
             \midrule
            NC    &  0.48 & 0.78$\pm 0.12$     \\
                    
            ABS  & 0.51 & 0.82$\pm 0.10$         \\
                 
            ULP     & 0.49  & 0.85$\pm 0.09$   \\
            AC    & 0.61 & 0.66$\pm 0.15$  
             \\
             
            IVA-RF (ours)    & \textbf{0.32} & \textbf{0.91$\mathbf{\pm 0.06}$}

            \\\bottomrule
        \end{tabular}
        
        \caption{Comparison of backdoor detection performance with four SOTA methods in image classification dataset. IVA-RF works better than others with low CE-Loss and high ROC-AUC.}\label{Tab2}
    \end{table}

\noindent\textit{\textbf{Object detection}}

The majority of backdoor attack detection techniques for image classification do not work for object detection. In addition, the object detection model's output (a large number of objects) differs from the image classification model (predicted class). The only SOTA method we have found to compare our algorithm with is detector cleanse (DC) \cite{chan2022baddet}   and the results are shown in Table 4. Similar to image classification, IVA-RF outperforms DC with higher ROC-AUC and lower CE-Loss.

\begin{table}[h!]
\setlength{\tabcolsep}{15pt}
   \begin{tabular}{lrr}\toprule
            & CE-Loss & ROC-AUC 
            \\  
             \midrule
            DC    &  0.48 & 0.81$\pm 0.12$     \\

            IVA-RF (ours)    & \textbf{0.41} & \textbf{0.89$\mathbf{\pm 0.09}$}

            \\\bottomrule
        \end{tabular}
        
        \caption{Comparison of backdoor detection performance with only comparable method available in object detection dataset and IVA-RF works better.}\label{Tab2}
    \end{table}

\subsubsection{Efficiency of the methods }

It's critical that backdoor detection techniques are effective because they may end up being a standard component of ML operations.  Table 5
shows the time in seconds required to make decisions for backdoor detection.  Our method  tends to be faster than NC, ABS, ULP (image classification), and DC (object detection) by an order
of magnitude due to the fact that our approach is
model agnostic and only extracts  features from model weights for  detection.  Although AC's running duration is close to ours, it is noticeably less accurate, as seen in Table 3. Because of this, our approach can achieve an efficiency-accuracy balance that none of the other algorithms can.

\begin{table}[h!]
{%

        \begin{tabular}{l|rrrrrr}\toprule
              &\multicolumn{6}{c}{{computation time of methods (s)}} 
              \\\cmidrule(r){2-7}
             Dataset&NC&ABS&ULP&AC&DC &{IVA-RF}  \\\midrule

             Image    & 1346  & 1565& 2514  & 267& - &\textbf{145}      \\
                    
            Object   & -  & -& -  & -& 23243  &\textbf{2164}      \\
            
              \hline

        \end{tabular}
        }
        \caption{Computation time in (s) including our algorithm: IVA-RF, and NC, ABS, ULP, AC, and DC.}\label{Tab2}
    \end{table}

\subsubsection{Ablation study}

As we have applied PCA for dimensionality reduction before IVA, an ablation  study was conducted to see the impact of PCA. Figure 4 shows the ROC-AUC scores when we do not use PCA and with different numbers of PCA components. The classifier performance degrades significantly when we do not use PCA as IVA has to handle the noisy data to extract features. However, we preserved 90\% variance of the data by using a number of components $N=4$ and 10 for image and object datasets respectively. When we use lower or higher numbers of components the score drops as we loose information for lower numbers and we add noisy components for higher numbers.

\begin{figure}[h!]
  
  \centering
  \includegraphics[width=2in,height=1.5in]{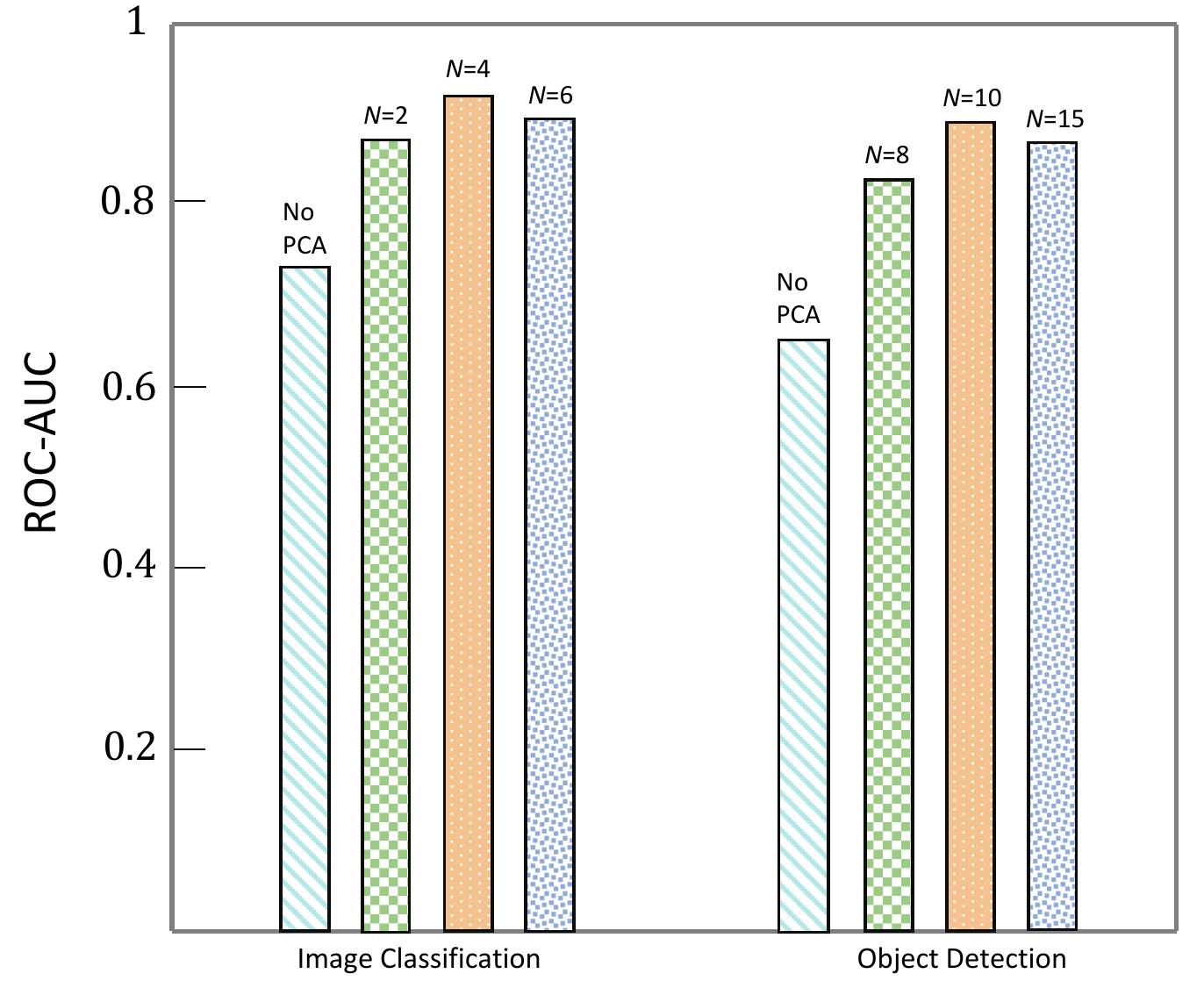}
  \caption{Impact of applying PCA and number of PCA components on the performance of our method.}
\end{figure}

\section{Conclusion}
Ours is the
first work of which we are aware that uses matrix factorization on the weights  to detect
backdoors in deep networks. Moreover, this is the first pipeline which can detect backdoor models in case of both image classification and object detection networks which has a number of advantages,
including the fact that it needs no re-training or optimization and is much
faster than other state-of-the-art backdoor detectors. 
Future work will include applications to sequence models
such as those used in natural language processing, which
should be straightforward from an engineering perspective
given that our method uses only the pre-trained weights of the networks.

\bibliography{sample-ceur}

\end{document}